\documentclass[5p,times,twocolumn,authoryear,nopreprintline]{elsarticle}
\usepackage{amsmath,amssymb}
\usepackage{booktabs}
\usepackage{multirow}
\usepackage{graphicx}
\usepackage{adjustbox}
\usepackage{placeins}
\usepackage{float}
\usepackage{xcolor}
\usepackage[hidelinks]{hyperref}

\newcommand{\alert}{\textsc{alert}}
\newcommand{\uncertain}{\textsc{uncertain}}
\newcommand{\noalert}{\textsc{no-alert}}
\newcommand{\op}[1]{\texttt{#1}}
\newcommand{\ind}{\mathbf{1}}
\newcommand{\yes}{$\checkmark$}
\newcommand{\no}{--}
\newcommand{\partly}{$\circ$}
\DeclareMathOperator*{\argmax}{arg\,max}
\graphicspath{{figures/}}
\makeatletter
\def\els@aparagraph[#1]#2{\elsparagraph[#1]{#2}}
\def\els@bparagraph#1{\elsparagraph*{#1}}
\makeatother

\begin{document}
\begin{frontmatter}
\title{SurgStreamer: proactive, request-driven monitoring of surgical video streams with structured state and a
reusable reasoning layer}

\author[vu]{Bowen Xiang\corref{cor1}}
\ead{bowen.xiang@vanderbilt.edu}
\author[vu]{Jon S. Heiselman}
\cortext[cor1]{Corresponding author.}
\affiliation[vu]{organization={Department of Biomedical Engineering, Vanderbilt University},
                 city={Nashville},
                 state={TN},
                 country={USA}}

\begin{abstract}
Surgical teams could benefit from assistants that watch the video for requested events, such as ``tell me when the
clip applier appears during Calot triangle dissection'', and answer questions about what has already happened. We
formalise this task as request-driven event detection with exact semantics and causal alert decisions, and present
SurgStreamer. It separates procedure-specific perception, which maps the 1-FPS stream to a structured state over
tools, actions and phases, from procedure-agnostic reasoning, which evaluates a formal monitoring condition over this
state and an event memory, either with exact rules or with a small controller that addresses entities through
pointers rather than names. We evaluate this monitoring core with gold conditions. On a pre-registered, sealed test
set of 12 laparoscopic cholecystectomies, SurgStreamer reached a macro event-F1 of 0.598 over full procedures. On
identical 75-s windows with only the window available to every method, it scored 0.677. The tested configurations of
the streaming vision--language model OneStreamer-4B, including ones fine-tuned on SurgStreamer's state or on video with
an adapted vision encoder, scored at most 0.540 (Holm-adjusted $p\le0.012$). At three external centres, re-training
only perception raised performance on held-out procedures from 0.239 to 0.470. On a new procedure, pituitary surgery,
controller weights trained on cholecystectomy matched an in-domain controller on oracle state (0.947 vs 0.938). On
predicted state they missed the pre-registered non-inferiority margin ($-0.024$ [$-0.040$, $-0.008$]) unless the
alert thresholds were re-calibrated. The results support separating perception from reasoning and point to
perception under distribution shift as the main limitation.
\end{abstract}

\end{frontmatter}

\section{Introduction}\label{sec:intro}
Laparoscopic cholecystectomy is among the most frequent abdominal operations. Its most feared complication, bile-duct
injury, is linked to the misidentification of structures before they are clipped and cut. Surgical guidelines
therefore require a critical view of safety before the cystic duct and artery are divided
\citep{strasberg2010cvs}, and recent work has automated the assessment of this view from video
\citep{mascagni2022cvs}. Beyond such fixed checks, a surgical team may want to ask an assistant to watch for events
of its choice: ``tell me when the cystic duct is being clipped'', ``alert me if the clip applier appears during Calot
triangle dissection'', or ``has the cystic artery been cut yet?''. Such requests are phrased in natural language but
refer to a predefined vocabulary of instruments, actions and phases, combined by a small set of operators. They are
time-critical: an answer is useful only if it arrives when the event happens, once per occurrence, and never for an
event that did not happen. In this work, events are visible events as defined by the dataset annotations; we do not
assess whether an action was performed correctly, the quality of structure handling, or the prevention of
complications.

\begin{figure*}[!tp]\centering
\includegraphics[width=\textwidth]{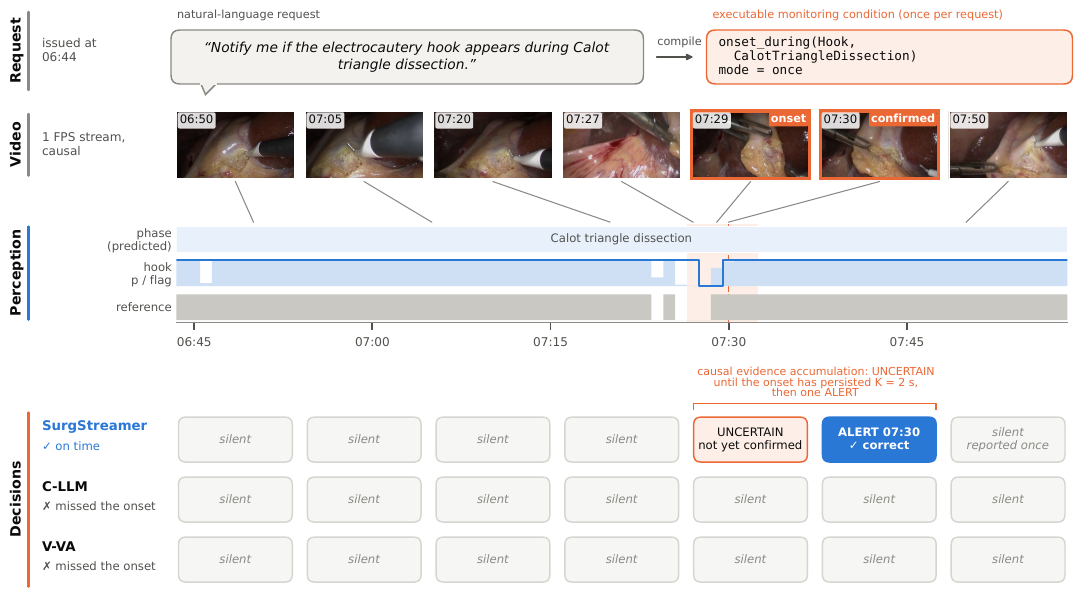}
\caption{Request-driven monitoring on a sealed Cholec80 test procedure (75-s window, equal context: every method has
only the window). The request is compiled once into an executable monitoring condition. From the 1-FPS frames,
perception computes a structured state: predicted phase, hook probability (shaded) and decoded flag (line); grey:
reference presence; orange band: matching tolerance around the reference occurrence; dashed line: its confirmation.
Bottom: per-frame decisions. SurgStreamer accumulates evidence causally: \uncertain{} at the onset second, a single
\alert{} once the onset has persisted $K=2$\,s, then silence. The strongest state-text baseline (C-LLM) and the
vision-adapted VLM (V-VA) stay silent over the whole window, as does V-LoRA (not shown). The example was selected for
illustration; Fig.~\ref{fig:failure} shows a failure, and Section~\ref{sec:results} the quantitative results.}
\label{fig:teaser}
\end{figure*}

Surgical data science has mainly addressed recognition \citep{maierhein2017sds}: phases
\citep{twinanda2017endonet,jin2018svrcnet,czempiel2020tecno,garrow2021review}, instruments, and actions
\citep{nwoye2022rendezvous}. These methods label frames or segments; they do not answer user requests, and their
metrics do not measure when an event is signalled. Streaming vision--language models (VLMs) can be prompted with such
requests \citep{chen2024videollmonline,onestreamer}, and streaming models now narrate surgery as it unfolds
\citep{he2026surgonair}. A recent time-aware benchmark found that general and medical VLMs degrade markedly when
they must decide \emph{when} to answer or alert \citep{wang2026medstreambench}. In the configurations we tested,
streaming VLMs alerted when an
instrument was merely visible rather than appearing, alerted on absent objects, rarely detected disappearances or
phase changes, and needed hundreds of milliseconds of GPU time per request and second of video. Fine-tuning reduced
but did not remove these failures, even when it included the vision encoder. Fig.~\ref{fig:teaser} shows a request on
a sealed test procedure, the decisions SurgStreamer takes on it, and the silence of the two strongest fine-tuned
VLMs.

We argue that proactive monitoring should be grounded in a structured state, and that the reasoning over this state
should be separated from the procedure. SurgStreamer (Fig.~\ref{fig:overview}) maps the video stream to a
probabilistic state over a closed vocabulary of tools, actions and phases. A formal monitoring condition is evaluated
over the state stream and an event memory, and alerts are decided causally. Perception is procedure-specific and must
be trained for each procedure, and possibly for each centre. Reasoning sees entities only through their kind and
state, never through their names, which makes it a candidate for reuse across procedures. A natural-language compiler
maps requests to conditions; it is an interface module, and our evaluation targets the monitoring core given formal
conditions. This paper tests three claims:
\begin{enumerate}
\item \textbf{Event-level monitoring with exact semantics.} Request-driven monitoring can be specified as event
detection with exact semantics and causal alert decisions, covering confirmation, once / every / $n$-th modes,
composition and retrospective memory. We provide a benchmark of 20 query families with exact references and hard
negatives.
\item \textbf{Name-free pointer control over structured state.} A small learned controller that reads the state
through entity pointers and structured temporal features improves event monitoring over the exact rules on a
pre-registered sealed test. Built on this state, the system outperforms the tested streaming-VLM configurations,
including fine-tuned and vision-adapted ones, under equal context.
\item \textbf{The transfer boundary.} Frozen controller weights transfer to a new procedure on oracle state, while
perception, and on predicted state the alert thresholds, must be adapted. We measure this at three external centres
and on a new procedure, and we report where the claim fails.
\end{enumerate}
The perception components (a fine-tuned DINOv3 encoder, causal temporal heads, an ordered phase model) are
established techniques that we use as a strong, documented substrate. Pre-registration and sealed test sets serve to
make the three claims testable.

\section{Related work}\label{sec:related}
\paragraph{Surgical workflow and instrument recognition.} Online phase recognition progressed from convolutional
features with recurrent models \citep{twinanda2017endonet,jin2018svrcnet} to temporal convolutions and transformers
\citep{czempiel2020tecno,gao2021transsvnet,liu2023skit,yang2024surgformer}; \citet{garrow2021review} review the
field. Instrument presence is usually learned jointly, and fine-grained actions are modelled as
instrument--verb--target triplets \citep{nwoye2022rendezvous}. Surgical foundation encoders pretrained on large video
collections aim at representations that transfer across tasks \citep{wang2023endofm}. Multi-centre benchmarks such
as HeiChole \citep{wagner2023heichole} and procedure-specific ones such as PitVis \citep{das2024pitvis} document
performance drops across centres and procedures. We use such recognition models as the perception layer and evaluate
event timing explicitly.

\paragraph{Online and event-level video understanding.} Online action detection \citep{degeest2016online} and action
spotting \citep{giancola2018soccernet} evaluate causal predictions and event timing within a tolerance. Our metric
follows this practice and adds request-specific semantics (modes, phase restrictions, ordering), a confirmation
interval defined from the annotations, and an explicit state for accumulating evidence.

\paragraph{Streaming and surgical vision--language models.} Streaming video LLMs decide at every frame whether to
respond \citep{chen2024videollmonline,wang2024mmduet,qian2025dispider}; some add a compressed long-term memory
\citep{zhang2024flashvstream}. We use OneStreamer-4B \citep{onestreamer}, a streaming fine-tune of Qwen3-VL
\citep{qwen3vl} with a native proactive protocol. Surgical VLMs address visual question answering and representation
learning \citep{seenivasan2022surgicalvqa,seenivasan2023surgicalgpt,yuan2023surgvlp}, mostly on frames or clips and
retrospectively. SurgOnAir \citep{he2026surgonair} brings streaming to surgery: it narrates the procedure frame by
frame and emits sparse state tokens at phase and step transitions, with steps generated by a language model from
narration transcripts. It describes the workflow rather than answering user requests, and it is evaluated by the
pairwise preferences of an LLM judge, not by the timing of the transitions it signals.

\paragraph{Time-aware evaluation.} MedStreamBench \citep{wang2026medstreambench} evaluates medical video
understanding under explicit evidence windows, across 22 datasets including Cholec80, HeiChole and PitVis. Its
proactive mode asks a model to answer \emph{no alert}, \emph{uncertain} or \emph{alert} as a stream unfolds, and
scores the first alert against the expected time (responsiveness) and whether the model stays positive afterwards
(stability). General and medical VLMs degraded markedly in its streaming and proactive settings. We adopt the same
three-way response space but evaluate it differently: an alert is an \emph{edge}, one per requested occurrence and
scored at every second against exact event definitions, so repeated alerts count as false alerts; and hard negatives
test whether a model stays silent when nothing new happens.

\paragraph{Structured reasoning and runtime monitoring.} Compositional visual questions can be answered by programs
over perception outputs \citep{gupta2023visprog,suris2023vipergpt}, and runtime verification monitors temporal-logic
properties over signals \citep{maler2004stl}. SurgStreamer combines these ideas for surgical streams and adds a
learned controller that approximates the exact semantics under perception noise. Table~\ref{tab:capability}
summarises the capabilities.

\begin{table}[!tbp]\centering\small
\caption{Capabilities of approach families, as typically published (\yes: yes, \partly: partly, \no: no).
Causal: decision at $t$ uses frames up to $t$. Events: evaluation of when events are signalled. Reuse: the
reasoning is reused across procedures without retraining.}
\label{tab:capability}
\begin{adjustbox}{max width=\linewidth}
\begin{tabular}{lccccc}\toprule
Approach & Causal & Request & Events & Memory & Reuse\\\midrule
Phase / tool recognition & \yes & \no & \no & \partly & \no\\
Surgical VQA / VLMs & \no & \yes & \no & \no & \partly\\
Streaming video LLMs & \yes & \yes & \partly & \partly & \yes\\
Surgical narration & \yes & \no & \no & \partly & \no\\
Program-based reasoning & \no & \yes & \no & \no & \yes\\
\textbf{SurgStreamer} & \yes & \yes & \yes & \yes & \partly\\
\bottomrule\end{tabular}
\end{adjustbox}
\end{table}

\section{Method}\label{sec:methods}

\begin{figure*}[!tp]\centering
\includegraphics[width=\textwidth]{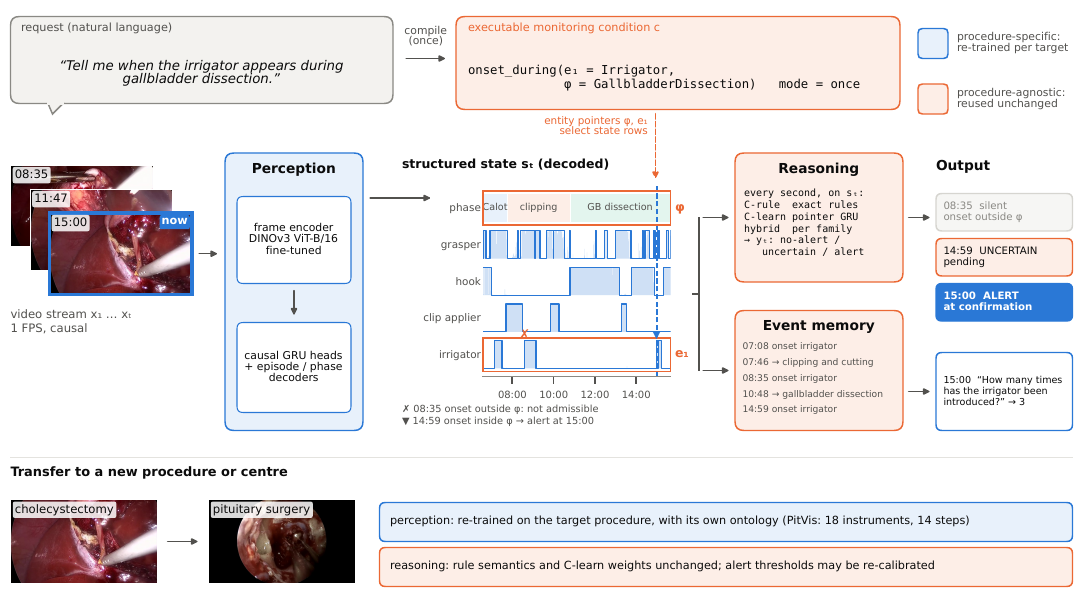}
\caption{SurgStreamer overview, shown on a sealed Cholec80 test procedure. A natural-language request is compiled once
into an executable monitoring condition. Procedure-specific perception (blue) turns the 1-FPS stream into a structured
state over a closed vocabulary of tools, actions and phases (decoded flags as lines, probabilities shaded; four of the
seven tools shown). Procedure-agnostic reasoning (orange) addresses the entities of the condition through pointers
($e_1$, $\varphi$), never by name. Every second, the exact rule controller C-rule, the learned controller C-learn or
their per-family hybrid outputs $y_t$, and an alert is issued causally, once the event is confirmed. Here the earlier
irrigator onset at 08:35 ($\times$) falls in the clipping phase and is not admissible; the onset at 14:59
($\blacktriangledown$) is confirmed and reported at 15:00 (dashed line), as in the reference (decisions of C-rule on the
decoded state). The event memory logs episodes and
phase transitions causally and answers retrospective requests; the count shown, 3, matches the reference. For a new
procedure or centre only the blue components are re-trained; the rule semantics and the C-learn weights are reused
unchanged, and only the alert thresholds may be re-calibrated.}
\label{fig:overview}
\end{figure*}

\subsection{Problem formulation}\label{sec:problem}
A video is observed as a causal stream of frames $x_1,x_2,\dots$ at 1 frame per second; time $t$ is in seconds. For
a \emph{proactive} request $r$ the system emits, at every second,
\begin{equation}\label{eq:policy}
y_t=\pi(r,x_{1:t})\in\mathcal{Y},
\end{equation}
with $\mathcal{Y}=\{\noalert,\uncertain,\alert\}$, the response space of the proactive mode of MedStreamBench
\citep{wang2026medstreambench}, for which we define each state exactly. One \alert{} must be issued per requested
occurrence, at the
moment the occurrence is confirmed, and the alert decision at $t$ may use only frames up to $t$. \uncertain{} marks
evidence that has started to accumulate; it is defined offline from the complete stream and serves as an auxiliary
training label (Section~\ref{sec:dsl}). A \emph{retrospective} request is answered at a question time $t_q$ from the
history. SurgStreamer
factorises $\pi$ into procedure-specific perception $f_\theta$, a request compiler $\Gamma$ and procedure-agnostic
reasoning $g$:
\begin{equation}\label{eq:factorisation}
\begin{gathered}
\mathbf{s}_t=f_\theta(x_{1:t}),\qquad c=\Gamma(r),\\
y_t=g(c,\mathbf{s}_{1:t}),
\end{gathered}
\end{equation}
where $\mathbf{s}_t$ is a probabilistic state over a vocabulary $\mathcal{E}$ and $c$ a monitoring condition. The
reasoning $g$ receives neither pixels nor entity names, only the state of the entities that $c$ refers to.
Fig.~\ref{fig:overview} shows the three parts on a test procedure: $f_\theta$ (blue), the condition $c$, and $g$ with
the event memory (orange).

\subsection{Entities, state and events}\label{sec:state}
The vocabulary is defined in a procedure ontology and contains entities of three kinds: instruments, actions (a verb
with an optional instrument and target) and phases. For cholecystectomy we use the 7 Cholec80 tools and phases
\citep{twinanda2017endonet}. We add 21 CholecT50 action classes \citep{nwoye2022rendezvous}: 9 verbs, 8
instrument--verb pairs, and the clipping and cutting of the cystic duct and artery. At second $t$, each tool or action
$e$ has a presence probability $p_e(t)$ and a binary flag $b_e(t)$ from an episode decoder (Section~\ref{sec:decoding});
phases have a posterior and a decoded phase $z_t$, and we write $b_\varphi(t)=\ind[z_t=\varphi]$.

Events are defined on a binary series $b$ observed from $t_s$, the first second of the stream (or of the window).
An \emph{episode} is a
maximal run of presence in which absences of at most $G=2$\,s are bridged. A \emph{candidate onset} starts at
\begin{equation}\label{eq:onsets}
\begin{aligned}
t_0\in\mathcal{O}_G(b)\iff\ &b(t_0)=1,\ t_0>t_s,\\
&b(t_0-j)=0\ \ \forall j\in\{1,\dots,\min(G+1,t_0-t_s)\}.
\end{aligned}
\end{equation}
The candidate is \emph{pending} on $[t_0,t_c)$ with $t_c=t_0+K-1$ and $K=2$. It is \emph{confirmed} at $t_c$ if the
entity remained present, $\prod_{j=0}^{K-1}b(t_0+j)=1$, and \emph{cancelled} at its first absent second otherwise. An
\emph{offset} is the last present second $t_\ell$ of an episode. It is pending while the entity is absent and confirmed
at $t_c=t_\ell+G+1$, once the absence has lasted $G+1$\,s; if the entity reappears earlier, the episode continues. A
phase entry is a candidate onset of $b_\varphi$ with $G=0$. Confirmation at $t_c$ depends on $b$ up to $t_c$ only, so
alert decisions are causal.

\subsection{Monitoring conditions}\label{sec:dsl}
A condition $c=(o,\mathbf{e},m,n,\varphi)$ consists of an operator $o$, entities $\mathbf{e}=(e_1[,e_2])$, a mode
$m\in\{$once, every, nth$\}$ with index $n$, and an optional phase restriction $\varphi$. Each proactive operator
defines a base series $\beta_c$ and an admissibility predicate $\alpha_c$ on its candidate onsets
(Table~\ref{tab:operators}). The mode selects the confirmed occurrences that are requested: the first, all, or the
$n$-th. The system alerts at the confirmation time $t_c$ of each selected occurrence and only then. A condition that
names an entity outside the vocabulary never alerts. In Fig.~\ref{fig:overview}, for example,
$c=(\op{onset\_during},\ \text{irrigator},\ \text{once},\ {-},\ \text{gallbladder dissection})$: the irrigator onset at 08:35
fails $\alpha_c$ because it occurs during clipping and cutting, and the first admissible onset (14:59) is reported at
its confirmation.

Per-second labels mark pending seconds \uncertain{} and confirmation seconds \alert{}. They are generated offline from
the complete stream and are not causal. A pending offset is known to be an offset only once it is confirmed, and in the
once mode a candidate that is later superseded is not labelled. These labels serve only as auxiliary training targets
for the learned controller (Section~\ref{sec:controllers}); evaluation uses alert times only. In a check on 2,000
random truncations of benchmark streams, the alerts on each prefix never differed from the alerts of the complete
stream, whereas 5 per-second labels did.

An exact evaluator $\mathrm{Eval}(c,\mathbf{s})$ implements these definitions. Applied to ground-truth state, it
defines the benchmark references. Applied to predicted state, it is the rule controller C-rule.

\begin{table}[!tbp]\centering\small
\caption{Proactive operators: base series $\beta_c$ and admissibility $\alpha_c$ of a candidate onset $t_0$.
Onset-after uses the whole procedure history.}
\label{tab:operators}
\begin{adjustbox}{max width=\linewidth}
\begin{tabular}{lll}\toprule
Operator & $\beta_c(t)$ & $\alpha_c(t_0)$\\\midrule
\op{onset}$(e)$ & $b_e(t)$ & 1\\
\op{phase\_enter}$(\varphi)$ & $b_\varphi(t)$, $G=0$ & 1\\
\op{co\_onset}$(e_1,e_2)$ & $b_{e_1}(t)\,b_{e_2}(t)$ & 1\\
\op{onset\_during}$(e,\varphi)$ & $b_e(t)$ & $\ind[z_{t_0}=\varphi]$\\
\op{onset\_after}$(e_1,e_2)$ & $b_{e_2}(t)$ & $\ind[\exists t'<t_0: b_{e_1}(t')=1]$\\
\op{offset}$(e)$ & \multicolumn{2}{l}{offsets of $b_e$, confirmed at $t_\ell+G+1$}\\
\bottomrule\end{tabular}
\end{adjustbox}
\end{table}

\subsection{Event memory}\label{sec:memory}
The event memory is an append-only log built causally from the decoded state. It logs each episode start at its
first second $t_0$, including an entity already present when the stream starts, and including starts that later fail
the $K$-second persistence. It logs each offset at its confirmation time and each phase transition at its first
second. A retrospective request at $t_q$ sees the events logged before $t_q$. \emph{Has} returns whether an episode of
the entity has started, \emph{count} the number of episode starts, and \emph{order} which of two entities started
first (or neither; a tie counts for the second). Counting therefore follows episode starts and is not restricted to confirmed onsets. Because the
log stores typed, time-stamped episodes rather than answers, one memory serves all retrospective requests about any
entity of the vocabulary; Figs.~\ref{fig:overview} and \ref{fig:memory} show it on test procedures.

\subsection{Request compiler (interface)}\label{sec:compiler}
The compiler is the natural-language interface to the monitoring core; the main evaluation uses gold conditions and
does not depend on it. A request $r$ is compiled by $\Gamma(r)=\mathcal{G}_r(\mathcal{N}(\mathrm{parse}(\mathrm{LM}(P,r))))$. Here
$\mathrm{LM}$ is the OneStreamer-4B language backbone used text-only with greedy decoding, prompted with the schema,
the vocabulary, five rules and one training example per family; $\mathrm{parse}$ validates the output against the
schema, and $\mathcal{N}$ normalises shape errors without guessing content. The lexical grounding $\mathcal{G}_r$ keeps
an entity only if one of its ontology stems occurs in $r$; otherwise the entity becomes unsupported and never alerts.
This turns the most consequential compilation error, mapping an absent object onto a similar real instrument, into
silence.

\subsection{Perception}\label{sec:perception}
Frames ($448\times256$) are encoded by DINOv3 ViT-B/16 \citep{dinov3,oquab2024dinov2}. The frame feature
$\mathbf{h}_t\in\mathbb{R}^{1536}$ concatenates the class token and the mean patch token. The last 8 of 12 blocks are
fine-tuned with linear heads for tools, phase and actions, using a loss that averages each label source over the
frames where it is available,
\begin{equation}\label{eq:frameloss}
\mathcal{L}_{\text{frame}}=\sum_{s}\frac{\sum_i m^{s}_i\,\ell_s(i)}{\max(1,\sum_i m^{s}_i)},
\end{equation}
where $i$ runs over the frames of a mini-batch, $m^{s}_i$ indicates whether frame $i$ has labels of source $s$, and $\ell_s$ is a class-weighted binary
cross-entropy (tools, actions; positive weight $\min(w_{\max},\max(1,(1-\mu_k)/\mu_k))$ for class prevalence $\mu_k$)
or a cross-entropy (phase). Training procedures are encoded by five out-of-fold encoders, so that downstream components
see realistic noise.

A causal temporal encoder, $\mathbf{v}_t=\mathrm{GELU}(\mathbf{W}\,\mathrm{LN}(\mathbf{h}_t)+\mathbf{d})$ followed
by a two-layer GRU \citep{cho2014gru}, feeds presence heads for $p_e(t)$ and the phase posterior. Optional event heads
detect onsets, offsets and phase entries within the first 3\,s from the event. The final state averages an ensemble of four
heads trained on 1024-s chunks; a three-head 256-s ensemble serves as an alternative state for the hybrid.

\subsection{Episode and phase decoding}\label{sec:decoding}
Flags are decoded causally per class by hysteresis. The flag switches on after $K_{\text{on}}$ consecutive seconds
with $p_e>T_{\text{on}}$, or when an onset head exceeds a threshold; it switches off after $K_{\text{off}}$ consecutive
seconds with $p_e<T_{\text{off}}$, or when an offset head exceeds a threshold. Phases are decoded by an ordered hidden Markov
model \citep{rabiner1989hmm}. Its transitions are estimated from training sequences, and transitions toward phases that
typically occur earlier are down-weighted:
\begin{equation}\label{eq:hmmA}
A_{ij}\propto\big(C_{ij}+\gamma\,\ind[i=j\vee C_{ij}>0]\big)\,\eta^{\ind[\rho_j<\rho_i]},
\end{equation}
where $C_{ij}$ are per-second transition counts in the training annotations (self-transitions included),
$\gamma=0.5$, $\rho_k$ is the mean relative position of phase $k$ in a procedure, and $\eta\le1$. With the phase
posterior $p^{\text{ph}}_k(t)$ of the temporal heads and the training frequency $\nu_k$ of phase $k$, the state is
filtered forward with tempered scaled likelihoods,
$\hat q_t(k)\propto(p^{\text{ph}}_k(t)/\nu_k)^{w}\sum_j\hat q_{t-1}(j)A_{jk}$, starting from
$\hat q_1(k)\propto\sigma_k(p^{\text{ph}}_k(1)/\nu_k)^{w}$ with $\sigma$ the training distribution of first phases
($\sigma=\nu$ for windows under equal context); the decoded phase switches only when the leading filtered
probability exceeds $\tau_{\text{ph}}$. The decoder parameters ($T_{\text{on}}$, $T_{\text{off}}$, $K_{\text{on}}$,
$K_{\text{off}}$, event-head thresholds, $\eta$, $w$, $\tau_{\text{ph}}$) are selected on the out-of-fold
predictions of all training procedures; inner five-fold cross-validation provides the reported development scores
and the decoded state of the training procedures themselves.

\subsection{Controllers}\label{sec:controllers}
C-rule applies the exact evaluator to the decoded state. C-learn reads the state of the entities named by the
condition through \emph{entity pointers} (the slots $e_1$ and $\varphi$ in Fig.~\ref{fig:overview}). For each of three slots (primary entity, secondary entity, phase
restriction) it computes eight causal features,
\begin{equation}\label{eq:slot}
\begin{aligned}
\boldsymbol{\psi}_t=\big[&p_t,\ b_t,\ u^{+}_t/10,\ u^{-}_t/10,\ N_t/5,\\
&\Delta^{\uparrow}_t/60,\ \Delta^{\downarrow}_t/60,\ \ind[\text{seen}]\big]_{[0,1]}.
\end{aligned}
\end{equation}
These are the probability and flag of the addressed entity, the lengths of the current run of present and of absent
seconds of the flag, the number of episode onsets so far, the times since the last episode onset and since the flag
last switched off, and whether the entity has been seen; $[\cdot]_{[0,1]}$ clips each feature to $[0,1]$. With an
11-dimensional query code (one-hot operator including ``unsupported'', an unsupported flag, one-hot kind of the
primary entity), the input has $3\times8+11=35$ dimensions. A linear layer with GELU, a 64-unit GRU and a linear
output map it to per-second probabilities $P_t$ over $\mathcal{Y}$.
Because C-learn never sees entity names, the same weights apply to unseen entity combinations and to other
vocabularies. Its design thus combines an entity-pointer interface with structured temporal features; we did not
separate the contributions of these features and of the recurrence.

C-learn is trained to detect every occurrence of the base event. The targets are the reference labels of the
condition with mode \emph{every} on ground-truth state, the inputs are the out-of-fold predicted state of the training
procedures (30\,\% ground-truth state), and the loss is a cross-entropy with class weights $\sqrt{N/N_y}$
($N_y$: number of training seconds with label $y$). At inference, alerts are rising edges of $P_t(\alert)$ above
per-operator thresholds $\tau_o$ (action and non-action operators calibrated separately on validation), at least
$G+1=3$\,s apart.
The requested mode is applied by structured decoding (first, every or $n$-th rising edge).

The hybrid chooses, for each query family $f$, a (state, controller) pair from the two state ensembles and the two
controllers:
\begin{equation}\label{eq:hybrid}
(s^{*}_f,\kappa^{*}_f)=\argmax_{(s,\kappa)}\mathrm{F1}_f(s,\kappa;\mathcal{P}_{\text{train}}).
\end{equation}
The selection was evaluated by nested cross-validation during development, then made once on all training folds and
frozen. Retrospective requests are answered by the event memory.

\section{Experimental setup}\label{sec:setup}
\subsection{Data and splits}\label{sec:data}
Cholec80 \citep{twinanda2017endonet} provides tool and phase labels, and CholecT50 \citep{nwoye2022rendezvous} action
labels, at 1\,FPS; together they contain 85 unique procedures (80 Cholec80, 5 CholecT50-only). The two datasets
annotate instrument presence under different protocols and are never mixed within a query. Splits respect the published
overlaps \citep{nwoye2022splits}. Before method development we sealed a final test set of 12 procedures (8 with action
labels). The remaining procedures form a training set (43, of which 27 have action labels), a validation set (9) and a
development test set (21) used for progress checks only. Procedure identifiers are listed in
\ref{app:splits}. Table~\ref{tab:datause} lists which data trained and selected each component.

HeiChole \citep{wagner2023heichole} contains 33 cholecystectomies from three German centres, with an official split
into procedures 1--24 and 25--33; for perception adaptation we used procedures 1--21 for training, 22--24 for
validation and 25--33 for testing. Its instrument categories map one-to-one onto five of our tools (grasper, clip
applier, scissors, irrigator, specimen bag), and its phases follow Cholec80. Queries about hook and bipolar, which
HeiChole pools into one category, are excluded. PitVis \citep{das2024pitvis} contains 25 endoscopic pituitary
operations with per-second labels of 18 instruments and 14 steps. For it we wrote a new ontology (instruments as
tools, steps as phases) and fixed a seeded 15 / 3 / 7 split; the 7 test procedures were sealed until evaluation.

\begin{table*}[!tp]\centering\small
\caption{Data used to train and to select each component (Cholec80 / CholecT50). OOF: out-of-fold predictions of the
training procedures. The 12 final-test procedures were used for nothing but the final evaluation.}
\label{tab:datause}
\begin{adjustbox}{max width=\linewidth}
\begin{tabular}{lll}\toprule
Component & Trained on & Selected / tuned on\\\midrule
Frame encoder & train (5 OOF models + 1 full) & val (checkpoint)\\
Temporal heads & train features & val (early stopping); train CV (ensemble)\\
Episode / phase decoders & -- & train OOF predictions (inner CV for scores)\\
C-learn & train OOF state (+30\,\% GT state) & val (checkpoint, thresholds)\\
Hybrid choices & -- & train CV (nested), then frozen\\
Request compiler & prompt from train templates & iterated on the development request set\\
LoRA baselines & train windows & val loss\\
\bottomrule\end{tabular}
\end{adjustbox}
\end{table*}

\subsection{Benchmark}\label{sec:benchmark}
References are computed by the exact evaluator on ground-truth state, using \emph{gold} monitoring conditions; all
main results use gold conditions, so they evaluate perception and reasoning, not the compiler. The 20 query families
(46 templates) cover tool onset and offset, phase entry, every and $n$-th occurrence, co-occurrence, onset during a
phase, onset after another event, action, instrument--verb and cystic duct / artery clip--cut events, absent objects,
and six retrospective families. Absent objects are instruments outside the vocabulary, so their silence is guaranteed
by construction; in-vocabulary objects that are simply not visible are not a separate family.

Each family has templates used only at test time. Two tool pairs and two instrument--verb pairs occur only in test
compositions: they are unseen by the controllers, although each entity is supervised in perception. In the
\emph{full-procedure} protocol every query is active over the whole procedure. The final test contains
$\approx$20,000 query instances. The \emph{windowed} protocol, needed to afford VLM baselines, uses 240 windows of
75\,s from 8 families: 7 scored by event-F1 and absent objects scored by false-positive rate. Half of the windows are
positive. Up to half of the negatives are \emph{persistent-presence} windows, in which the requested entity is visible
throughout but nothing new happens. The benchmark thus complements MedStreamBench \citep{wang2026medstreambench}:
instead of a few question rounds per stream, it scores a decision at every second, derives every reference from the
exact semantics of Section~\ref{sec:dsl}, and pairs positives with negatives that require silence.

\subsection{Baselines}\label{sec:baselines}
All VLM baselines use OneStreamer-4B \citep{onestreamer} with its native proactive protocol (silence, standby and
response tokens mapped to $\mathcal{Y}$). They use its conservative three-state system prompt, 1-FPS input with frames
capped at $224^2$ pixels, and a sliding window of 16 updates. Each update consists of a frame and the model's reply,
and the window drops frames and replies together, so no dialogue history survives beyond 16\,s. No additional memory
module was used. Decoding used temperature 0.7, top-$p$ 0.8 and top-$k$ 20, with one run at seed 42. Our VLM
conclusions are limited to this configuration.

The zero-shot baselines are \textbf{V} (video), \textbf{P-SG} (SurgStreamer's state rendered as text, with a prompt
that treats it as ground truth) and \textbf{VP-SG} (video plus state text). The fine-tuned baselines are \textbf{C-LLM}
(P-SG with LoRA \citep{hu2022lora}), \textbf{V-LoRA} (V with LoRA on the language model) and \textbf{V-VA}, a
vision-adapted model. V-VA adds LoRA on all 24 vision-encoder blocks and fully trains the vision--language projectors
(109\,M parameters). Half of its training updates describe the current surgical state, built from the same tool, phase
and action labels that train SurgStreamer's perception. All fine-tuned baselines are trained on the same 43 procedures,
with teacher forcing through the unchanged inference session, so that training and inference contexts are identical.

\subsection{Protocols and pre-registration}\label{sec:protocols}
In the original window protocol, SurgStreamer's state was computed from the whole procedure, while the VLMs saw the
window. In the \emph{equal-context} protocol, SurgStreamer is reset at each window start: its temporal heads start
from a zero hidden state, its decoders start cold, its phase model starts from class priors, and its controller
features start at the window. The state-reading baselines then read this reset state. Equal context equalises the
\emph{available} history. How much of it a method can \emph{use} still differs: SurgStreamer accumulates a state over
the 75\,s, while OneStreamer attends to its last 16 updates. A whole-procedure window reproduces the full-history
state exactly.

For transfer we distinguish three levels: the rule semantics (DSL definitions and the constants $G$ and $K$) are
unchanged; the controller weights are frozen; and the alert thresholds are either frozen or, as a separate condition,
re-calibrated on target validation data. On HeiChole we evaluate the whole system without retraining and then
re-train only perception with the same recipe, keeping the controllers, thresholds and hybrid choices fixed. On PitVis
we re-train perception on the 15 training procedures and compare frozen C-learn weights, with frozen and with
re-calibrated thresholds, against a C-learn trained in-domain, on oracle state (the ground-truth state) and on predicted state. An exploratory
few-shot analysis (\ref{app:fewshot}) trains a light temporal head on $k\in\{1,3,5,10\}$ annotated target
procedures.

Table~\ref{tab:prereg} gives the timeline. Each protocol was committed, with a time stamp, to the project's private
version-controlled repository before the corresponding data were scored. The protocols were not registered on a
public registry; the documents and their version history will be released with the code.

\begin{table}[!tbp]\centering\small
\caption{Protocol timeline (UTC, commit time in the private repository). The original window comparison with the VLMs
was part of the final protocol; the equal-context comparison and V-VA were defined after the final test had been
opened and are post hoc analyses of that test, but were fixed before the HeiChole windows were scored. $\delta$: matching
tolerance (Section~\ref{sec:metrics}).}
\label{tab:prereg}
\begin{adjustbox}{max width=\linewidth}
\begin{tabular}{p{0.60\linewidth}l}\toprule
Protocol (scope) & Committed\\\midrule
Final protocol: frozen system, hybrid primary, macro event-F1 ($\delta=2$\,s tool / action, 10\,s phase$^{a}$),
paired bootstrap, secondary metrics, baselines V, P-SG, VP-SG, C-LLM, V-LoRA & 2026-10-07 06:16\\
\quad final test first opened & 2026-10-07 13:46\\
Equal-context windows; V-VA (post hoc for the final test) & 2026-10-08 04:35\\
HeiChole external test; PitVis transfer with non-inferiority margin 0.03 and threshold re-calibration (secondary) &
2026-10-08 05:09\\
HeiChole perception-only adaptation & 2026-10-08 09:11\\
Few-shot pools and adoption rules (test curves exploratory) & 2026-10-09 05:44\\
\bottomrule\end{tabular}
\end{adjustbox}
\par\smallskip\footnotesize $^{a}$Chosen on development data before the final protocol. Reporting choices added later:
add-one $p$-value correction and Holm adjustment.
\end{table}

\subsection{Metrics and statistics}\label{sec:metrics}
An alert at $a$ matches an occurrence $(t_e,t_c)$, with event time $t_e=t_0$ for onsets and $t_e=t_\ell$ for
offsets, if $t_e-\delta\le a\le t_c+\delta$, greedily in time and one-to-one. For a family $f$, with $\mathrm{TP}_f$
matched occurrences, $N^{\text{occ}}_f$ occurrences and $N^{\text{al}}_f$ alerts, each summed over all requests and
procedures of the family,
\begin{equation}\label{eq:eventf1}
\mathrm{F1}_f=\frac{2\,\mathrm{TP}_f}{N^{\text{occ}}_f+N^{\text{al}}_f},\qquad
\overline{\mathrm{F1}}=\frac{1}{|\mathcal{F}|}\sum_{f\in\mathcal{F}}\mathrm{F1}_f.
\end{equation}
The macro event-F1 $\overline{\mathrm{F1}}$ averages over the set $\mathcal{F}$ of proactive families scored by
event-F1 (absent objects excluded); a family with $N^{\text{occ}}_f+N^{\text{al}}_f=0$ in a (resampled) set is
omitted. Negatives are scored by
their false-positive rate (FPR), and retrospective requests by balanced accuracy (has, order) and exact / within-one
accuracy (count). We also report false alerts per monitored request-hour and the alert delay after the event, which
includes the confirmation interval and decoder hysteresis but not computation time.

Uncertainty is quantified by a procedure-level bootstrap ($B=1000$) with percentile intervals
\citep{efron1993bootstrap}. Differences are paired, and controller seeds are averaged within each resample. Two-sided
$p$-values use the add-one correction $p=\min\{1,2(1+n)/(1+B)\}$, where $n$ is the number of resampled
differences in the smaller tail ($\le0$ or $\ge0$), so the smallest attainable value is $2/1001\approx0.002$. Within each family of comparisons against several baselines, they are
Holm-adjusted. Full training and implementation details are provided in \ref{app:impl}.

\section{Results}\label{sec:results}

\subsection{Sealed final test}\label{sec:res_final}

\begin{table}[!tbp]\centering\small
\caption{Final test (12 sealed procedures), full-procedure protocol: event-F1 per family (mean over seeds) and macro event-F1; C-learn and C-rule on the 1024-s state. 95\,\% procedure-level bootstrap intervals of the macro event-F1: Hybrid [0.550, 0.651]; C-learn (1024-s) [0.559, 0.661]; C-rule (1024-s) [0.508, 0.632]. Absent-object requests: no alert in any procedure for all three systems.}\label{tab:final_families}
\begin{adjustbox}{max width=\linewidth}
\begin{tabular}{lccc}\toprule
Family & Hybrid & C-learn & C-rule \\\midrule
Tool onset & 0.84 & 0.84 & 0.85 \\
Tool offset & 0.66 & 0.66 & 0.70 \\
Phase entry & 0.66 & 0.66 & 0.58 \\
Tool, every & 0.73 & 0.73 & 0.64 \\
Tool, $n$-th & 0.58 & 0.56 & 0.52 \\
Co-occurrence & 0.67 & 0.73 & 0.67 \\
Onset during phase & 0.64 & 0.64 & 0.48 \\
Onset after event & 0.71 & 0.73 & 0.71 \\
Action onset & 0.45 & 0.45 & 0.45 \\
Action, every & 0.61 & 0.65 & 0.59 \\
Instrument--verb onset & 0.35 & 0.42 & 0.38 \\
Action during phase & 0.45 & 0.44 & 0.45 \\
Duct / artery clip--cut & 0.41 & 0.37 & 0.38 \\
\midrule Macro event-F1 & 0.598 & 0.607 & 0.570 \\
\bottomrule\end{tabular}
\end{adjustbox}\end{table}

\begin{table*}[!tp]\centering\small
\caption{Final test, full procedures (pre-registered secondary metrics). Registered: macro event-F1 with $\delta=2$\,s for tool / action events and 10\,s for phase entry. $\delta=d$: the same tolerance for all families. FA/h: unmatched alerts per monitored request-hour. Delay: median / 90th percentile of the alert time after the event (s), matched alerts; it includes the confirmation interval and decoder hysteresis, not computation ($\approx$10\,ms).}\label{tab:monitoring}
\begin{adjustbox}{max width=\linewidth}
\begin{tabular}{lccccccc}\toprule
System & Registered & $\delta=2$ & $\delta=5$ & $\delta=10$ & $\delta=20$ & FA/h & Delay \\\midrule
Hybrid & 0.598 & 0.575 & 0.627 & 0.663 & 0.717 & 0.67 & 1 / 2 \\
C-learn (1024-s) & 0.607 & 0.584 & 0.615 & 0.644 & 0.688 & 0.62 & 1 / 2 \\
C-rule (1024-s) & 0.570 & 0.529 & 0.597 & 0.650 & 0.697 & 0.77 & 1 / 3 \\
\bottomrule\end{tabular}
\end{adjustbox}\end{table*}

On the 12 sealed procedures, the pre-registered primary system, the hybrid, reached a macro event-F1 of 0.598 [0.550,
0.651] with gold conditions (Table~\ref{tab:final_families}). This is consistent with cross-validation (0.579) and the
development test (0.563). It exceeded the exact rules on both state ensembles: by +0.028 [+0.009, +0.045] against
C-rule on the 1024-s state, and by +0.036 [+0.022, +0.054] on the 256-s state (both Holm-adjusted $p=0.006$; on the
development test the corresponding margin was +0.037). C-learn alone scored nominally higher (0.607) but was not
significantly different from the hybrid ($-0.009$ [$-0.025$, +0.007]). We therefore do not claim that the hybrid is
the best variant.

Under a strict tolerance of 2\,s for every family, including phase entry, the scores were 0.575 (hybrid), 0.584
(C-learn) and 0.529 (C-rule), the same order as under the registered tolerances (Table~\ref{tab:monitoring}). Among
the alerts that matched an occurrence, the median delay after the event was 1\,s and the 90th percentile 2\,s,
consistent with the 1-s confirmation interval of onsets (offsets need $G+1=3$\,s); this does not imply that every event was detected within 2\,s. The
hybrid produced 0.67 false alerts per monitored request-hour, that is, per hour for each active request; the alert
burden of a whole operation grows with the number of active requests.

\begin{figure*}[!tp]\centering
\includegraphics[width=\textwidth]{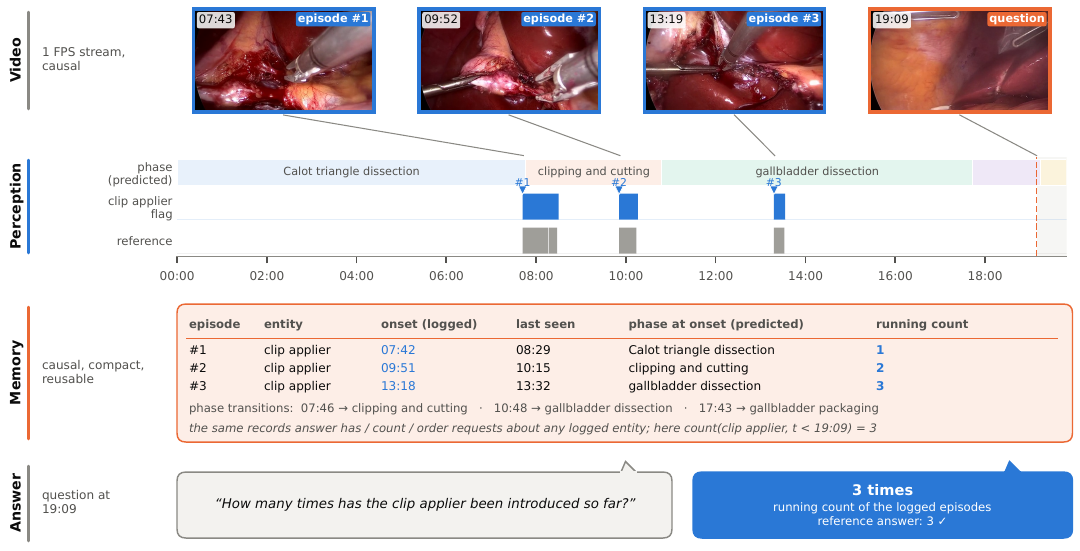}
\caption{Event memory on a sealed Cholec80 test procedure (full history). Perception decodes the clip applier (flag,
blue; reference, grey) and the phase. The memory logs, causally, one record per episode (onset, last-seen time,
predicted phase at onset) and the phase transitions. At 19:09 the question ``How many times has the clip applier been
introduced so far?'' is answered by the running count of the logged episodes (3, as in the reference); the logged
onsets coincide with the reference onsets. The same records answer has / count / order requests about any logged
entity. Selected for illustration.}
\label{fig:memory}
\end{figure*}

Per family, tool onsets were detected with F1 0.84, phase entries with 0.66, relational requests with 0.64--0.71 and
action onsets with 0.45. No out-of-vocabulary request triggered an alert, as guaranteed by construction. The event
memory answered occurrence and order questions with balanced accuracy of at least 0.91 and counted within one episode
in 85--86\,\% of cases (Table~\ref{tab:memory}). Counting was the hardest retrospective task (exact accuracy 0.60 for
tools): the count follows the decoded episodes, so every spurious or fragmented episode changes it; Fig.~\ref{fig:memory}
shows a count answered correctly. Requests whose entity composition never occurs in training remained hard for every
controller (0.40--0.48; Section~\ref{sec:res_dev}).

\begin{table}[!tbp]\centering\small
\caption{Final test: retrospective requests answered from the event memory (balanced accuracy for has / order; exact accuracy, within-one accuracy and mean absolute error for counts).}\label{tab:memory}
\begin{adjustbox}{max width=\linewidth}
\begin{tabular}{lcc}\toprule
 & Tools & Actions \\\midrule
Has occurred (bal. acc.) & 0.995 & 0.974 \\
Which first (bal. acc.) & 0.994 & 0.915 \\
Count, exact & 0.601 & 0.515 \\
Count, within one & 0.863 & 0.848 \\
Count, MAE & 0.68 & 0.70 \\
\bottomrule\end{tabular}
\end{adjustbox}\end{table}

\subsection{Comparison with streaming vision--language models}\label{sec:res_vlm}

\begin{table}[!tbp]\centering\small
\caption{Cost per 1-s update on a single RTX 5070 (12\,GB). The frame encoder and temporal heads run once per stream, independent of the number of requests; C-learn processes concurrent requests as one batch. The request compiler runs once when a request is issued and is excluded.}\label{tab:efficiency}
\begin{adjustbox}{max width=\linewidth}
\begin{tabular}{lr}\toprule
Component & Latency \\\midrule
DINOv3 frame encoder & 8.1 ms \\
Temporal heads (4 members) & 1.4 ms \\
C-learn, 1 / 10 / 100 requests & 0.15 / 0.16 / 0.18 ms \\
Exact DSL evaluator & 0.38 $\mu$s / request-s \\
Peak GPU memory & 0.64 GB \\\midrule
OneStreamer-4B, one request & 340--740 ms, $\approx$9 GB \\
\bottomrule\end{tabular}
\end{adjustbox}\end{table}

\begin{table*}[!tp]\centering\small
\caption{Comparison with streaming vision--language models (OneStreamer-4B, 16-update window) on identical 75-s windows, gold conditions. Macro event-F1 with 95\,\% procedure-bootstrap intervals over the families scored by event-F1 (Cholec80: 7 of 8 window families; absent objects are scored by false-positive rate. HeiChole: 4 primary families). \emph{Equal}: every method has only the window available (SurgStreamer re-computes its state from the window start); \emph{full}: the state read by the state-based methods was computed from the whole procedure (the pre-registered protocol). $\Delta$: SurgStreamer (equal) minus the method; $p_{\text{H}}$: paired bootstrap, add-one corrected, Holm-adjusted over the six baselines of each dataset. FPR$_{\text{pp}}$: share of persistent-presence windows with an alert (equal context). HeiChole: three external centres, all models trained on Cholec80 / CholecT50 only.}\label{tab:windows}
\begin{adjustbox}{max width=\linewidth}
\begin{tabular}{llcccccccc}\toprule
 & & \multicolumn{5}{c}{Cholec80 final test (12 procedures)} & \multicolumn{3}{c}{HeiChole (33 procedures)} \\\cmidrule(lr){3-7}\cmidrule(lr){8-10}
Method & Input & Full & Equal & $\Delta$ & $p_{\text{H}}$ & FPR$_{\text{pp}}$ & Equal & $\Delta$ & $p_{\text{H}}$ \\\midrule
\textbf{SurgStreamer (ours)} & state & 0.685 & 0.677 [0.577, 0.759] & -- & -- & 0.16 & 0.363 [0.252, 0.457] & -- & -- \\
C-LLM & state text & 0.557 & 0.540 [0.427, 0.638] & +0.137 & 0.012 & 0.02 & 0.155 [0.067, 0.239] & +0.208 & 0.036 \\
P-SG & state text & 0.430 & 0.449 [0.339, 0.546] & +0.228 & 0.012 & 0.69 & 0.234 [0.127, 0.336] & +0.128 & 0.104 \\
VP-SG & video + state & 0.307 & 0.340 [0.256, 0.435] & +0.337 & 0.012 & 0.84 & 0.221 [0.131, 0.300] & +0.141 & 0.070 \\
V-VA & video & -- & 0.473 [0.390, 0.545] & +0.204 & 0.012 & 0.07 & 0.356 [0.236, 0.477] & +0.007 & 0.969 \\
V-LoRA & video & -- & 0.377 [0.276, 0.470] & +0.300 & 0.012 & 0.16 & 0.446 [0.300, 0.577] & -0.083 & 0.723 \\
V (zero-shot) & video & -- & 0.116 [0.069, 0.165] & +0.561 & 0.012 & 0.89 & 0.182 [0.085, 0.287] & +0.180 & 0.104 \\
\bottomrule\end{tabular}
\end{adjustbox}\end{table*}

\begin{figure*}[!tp]\centering
\includegraphics[width=\textwidth]{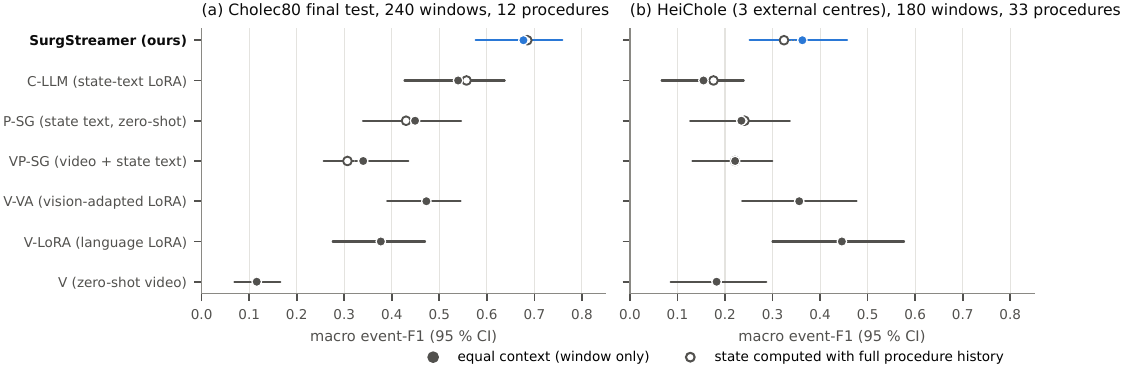}
\caption{Comparison with streaming VLMs under equal context (gold conditions). Macro event-F1 with 95\,\%
procedure-bootstrap intervals on identical 75-s windows. Filled dots: every method has only the window available.
Open dots: the state given to state-reading methods was computed with the full procedure history (pre-registered
protocol). (a) Cholec80 final test. (b) HeiChole, three external centres, all models trained on Cholec80 / CholecT50 only.}
\label{fig:windows}
\end{figure*}

In the pre-registered protocol, in which SurgStreamer's state was computed with full history, the hybrid (0.685)
exceeded every tested OneStreamer configuration on the final windows, at a fraction of their computational
cost (Table~\ref{tab:efficiency}). The difference was +0.128 against C-LLM and
+0.569 against zero-shot video (Holm-adjusted $p\le0.010$). In the post hoc equal-context analysis the conclusion was
unchanged (Table~\ref{tab:windows}, Fig.~\ref{fig:windows}a). SurgStreamer scored 0.677 [0.577, 0.759], and its
margins ranged from +0.137 over C-LLM, the strongest baseline, and +0.204 over the vision-adapted V-VA, to +0.561 over
zero-shot video (Holm-adjusted $p\le0.012$ for all six). Removing SurgStreamer's own history changed its score by only
$-0.008$. SurgStreamer scored highest at every matching tolerance from 2 to 20\,s, although the order among the
baselines changed (Table~\ref{tab:tolerance_windows}).

\begin{figure*}[!tp]\centering
\includegraphics[width=\textwidth]{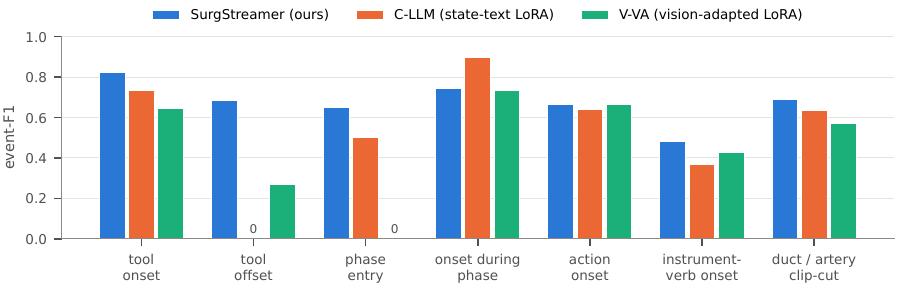}
\caption{Event-F1 per family on the Cholec80 final-test windows under equal context. The VLMs detect appearances but
miss disappearances and phase entries.}
\label{fig:families}
\end{figure*}

\begin{figure*}[!tp]\centering
\includegraphics[width=\textwidth]{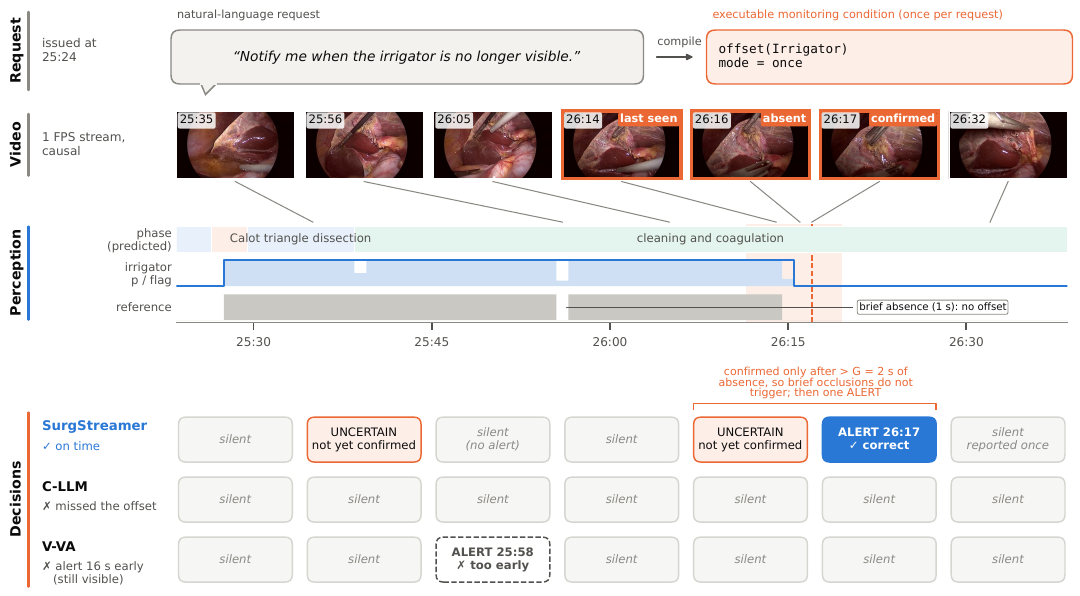}
\caption{A disappearance request on a sealed Cholec80 test window (equal context). The event is the last second the
irrigator is visible (``last seen''); it is confirmed only after the instrument has been absent for more than $G=2$\,s,
so the 1-s gap in the reference annotation (annotated) is not an offset. C-learn, which answers this family, briefly
outputs \uncertain{} around a dip of the irrigator probability without alerting, and alerts once at the confirmation of
the real disappearance. C-LLM stays silent, as does V-LoRA (not shown); V-VA alerts 16\,s early, while the irrigator
is still visible. Selected to illustrate the failure mode quantified in Fig.~\ref{fig:families}; layout as in
Fig.~\ref{fig:teaser}.}
\label{fig:offset}
\end{figure*}

Adapting the VLM's visual pathway improved video-only performance by 0.096 (V-VA over V-LoRA) but left the
characteristic failure intact. V-VA detected appearances well but missed disappearances (tool-offset F1 0.27 vs 0.69)
and phase entries (0.00 vs 0.65); both require state over time (Fig.~\ref{fig:families}). Fig.~\ref{fig:offset} shows
a disappearance request: SurgStreamer waits until the absence has lasted longer than $G$, which also keeps a 1-s gap
in the reference from counting as an offset, whereas V-VA alerts 16\,s early, while the instrument is still visible,
and C-LLM stays silent. The VLMs behaved differently on negatives. Zero-shot video alerted on all absent-object windows and on 89\,\% of
persistent-presence windows. C-LLM and V-VA were rarely triggered by persistent presence (FPR 0.02 and 0.07), but they
achieved this partly by staying silent, which lowered their recall. SurgStreamer's persistent-presence FPR was 0.16. For families answered by the exact rules, such
alerts require an error of the decoded state, such as a visible instrument decoded as absent for at least 3\,s or a
spurious phase switch; for families answered by the learned controller, a controller error is also possible.

Excluding the request compiler, which runs once when a request is issued, SurgStreamer needs about 10\,ms and 0.64\,GB
of GPU memory per 1-s update (Table~\ref{tab:efficiency}). Most of the time is spent on the frame encoder and temporal
heads, which are shared by all requests. The batched controller took 0.15, 0.16 and 0.18\,ms for 1, 10 and 100
concurrent requests. The tested OneStreamer configuration needed 0.34--0.74\,s and $\approx$9\,GB per update for one
request; additional requests need their own streaming sessions and forward passes, although the model weights could
be shared. For a single request, SurgStreamer is therefore 34--74 times faster and needs 14 times less GPU memory.

\subsection{External centres}\label{sec:res_heichole}

\begin{table}[!tbp]\centering\small
\caption{External centres (HeiChole), full-procedure protocol, 8 tool / phase families. Without retraining: the whole Cholec80-trained system. Adapted: perception re-trained with the same recipe on HeiChole procedures 1--21 (validation 22--24), with controller weights, thresholds and hybrid choices unchanged. Adapted $-$ without retraining (hybrid, procedures 25--33): +0.231 [+0.190, +0.277], $p$ $\le$0.002 (paired bootstrap, add-one corrected).}\label{tab:heichole}
\begin{adjustbox}{max width=\linewidth}
\begin{tabular}{lccc}\toprule
 & No retraining & No retraining & Adapted perception \\
Controller & (1--33) & (25--33) & (25--33) \\\midrule
Hybrid & 0.231 & 0.239 & 0.470 [0.400, 0.547] \\
C-learn (1024-s) & 0.226 & 0.235 & 0.449 [0.391, 0.507] \\
C-rule (1024-s) & 0.213 & 0.210 & 0.495 [0.428, 0.560] \\
\bottomrule\end{tabular}
\end{adjustbox}\end{table}

Applied to all 33 HeiChole procedures without retraining, the system scored 0.231 [0.202, 0.258]
(Table~\ref{tab:heichole}). On HeiChole's ground-truth state the rules reproduced every reference; this confirms that
the benchmark construction is consistent, but it does not by itself exclude errors of the learned controller or of
the decoders. The diagnostics point to perception and event timing as the main limitation. Frame-level recognition
remained high for grasper, irrigator and specimen bag (average precision 0.96, 0.91 and 0.86) but dropped for clip
applier and scissors (0.40 and 0.25). Grasper episodes fragmented (59 predicted vs 38 true per procedure). In
addition, HeiChole marks an instrument present from the first appearance of its tip, which shifts onsets: grasper
onset F1 was 0.38 at $\pm$2\,s and 0.64 at $\pm$30\,s. Re-training only perception on procedures 1--21, with controllers, thresholds and hybrid choices unchanged, raised
performance on procedures 25--33 from 0.239 to 0.470 (+0.231 [+0.190, +0.277], $p\le0.002$). On this adapted state
the learned controllers were not better than the exact rules.

On the HeiChole windows with all models trained on Cholec80 only (Table~\ref{tab:windows}, Fig.~\ref{fig:windows}b),
SurgStreamer scored 0.363. After Holm adjustment its advantage was significant only against C-LLM (+0.208, $p=0.036$).
The raw-video V-LoRA was nominally better (0.446, n.s.).

\subsection{A new procedure: controller transfer to PitVis}\label{sec:res_pitvis}

\begin{table}[!tbp]\centering\small
\caption{New procedure (PitVis, sealed test, 7 procedures, evaluated once); rows are C-learn unless named. Transferred: controller weights trained on cholecystectomy and reused unchanged, with the cholecystectomy thresholds unless re-calibrated on 3 PitVis validation procedures; on oracle and on predicted state (perception re-trained on 15 PitVis procedures). Transferred $-$ in-domain C-learn on predicted state: -0.024 [-0.040, -0.008]; on oracle state: +0.009 [-0.012, +0.029].}\label{tab:pitvis}
\begin{adjustbox}{max width=\linewidth}
\begin{tabular}{lcc}\toprule
Controller & Oracle state & Predicted state \\\midrule
Transferred & 0.947 [0.928, 0.965] & 0.372 [0.299, 0.452] \\
Transferred, re-calibrated & -- & 0.397 [0.326, 0.477] \\
In-domain (PitVis) & 0.938 [0.928, 0.946] & 0.397 [0.331, 0.468] \\
Transferred hybrid & -- & 0.376 [0.309, 0.451] \\
C-rule (exact) & 1.000 & 0.355 [0.284, 0.435] \\
\bottomrule\end{tabular}
\end{adjustbox}\end{table}

\begin{figure}[!tbp]\centering
\includegraphics[width=\linewidth]{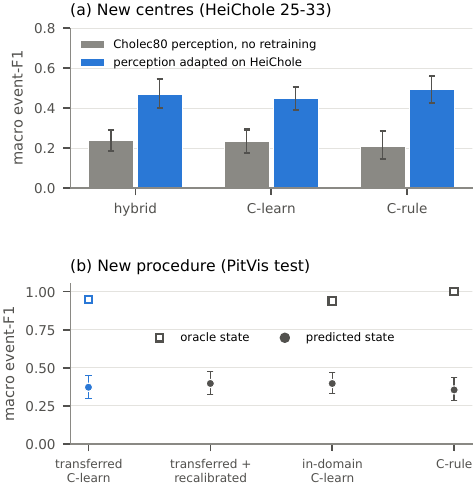}
\caption{Transfer with frozen controller weights. (a) New centres (HeiChole 25--33): re-training only the perception
doubles performance, with controllers, thresholds and hybrid choices unchanged. (b) New procedure (PitVis sealed test):
oracle state (squares) and predicted state (dots); ``re-calibrated'' keeps the weights frozen and re-calibrates only
the alert thresholds. C-rule is exact on oracle state by construction.}
\label{fig:transfer}
\end{figure}

On the sealed PitVis test (Table~\ref{tab:pitvis}, Fig.~\ref{fig:transfer}b), the C-learn weights trained on
cholecystectomy scored 0.947 [0.928, 0.965] on oracle state, and a C-learn trained on PitVis scored 0.938 (difference
+0.009 [$-0.012$, +0.029]). Both remain below the exact rules, which are correct on oracle state by construction
(1.000). The result therefore shows that the learned controller is independent of the procedure, not that it is
superior.

\begin{figure*}[!tp]\centering
\includegraphics[width=\textwidth]{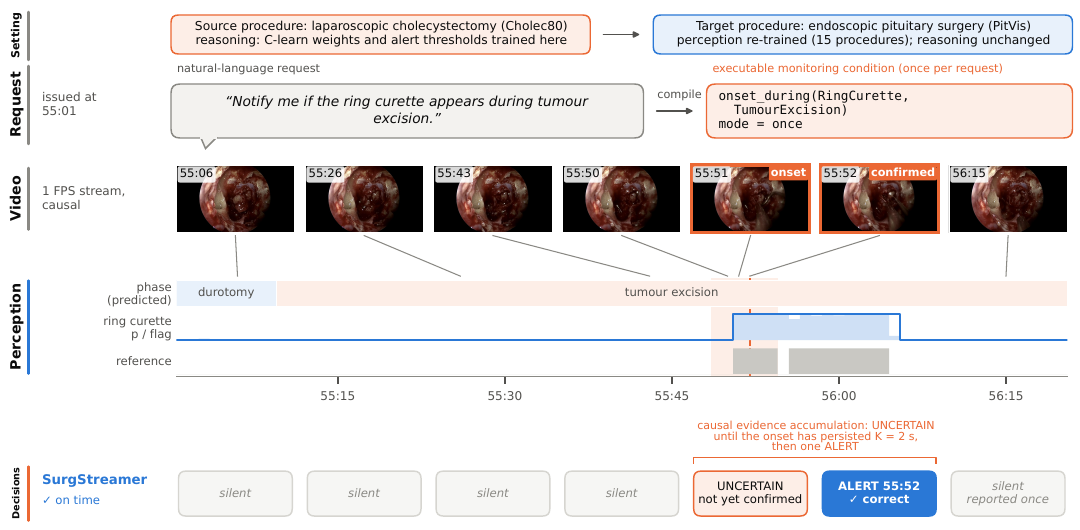}
\caption{The reasoning trained on cholecystectomy, applied unchanged to pituitary surgery (PitVis sealed test). Top:
the transfer setting. Perception was re-trained on 15 PitVis procedures; the C-learn weights and alert thresholds are
those trained on Cholec80. The controller addresses the ring curette and the tumour-excision phase only through
entity pointers; it outputs \uncertain{} at the onset and alerts once at confirmation. Selected for illustration; over
the whole test set, the transferred controller scored 0.372 on predicted state (Table~\ref{tab:pitvis}). Layout as in
Fig.~\ref{fig:teaser}.}
\label{fig:pitvis_example}
\end{figure*}

On predicted state, with perception re-trained on PitVis (tool onset / offset F1 $\approx$0.60 in training
cross-validation), the two transfer conditions differed. With frozen weights and frozen thresholds, C-learn scored
0.372. This exceeded the exact rules (+0.018 [+0.004, +0.031]) but stayed below the in-domain controller ($-0.024$
[$-0.040$, $-0.008$]), so the pre-registered non-inferiority criterion (margin 0.03) was not met. With frozen weights
and thresholds re-calibrated on three validation procedures (the pre-registered secondary analysis), C-learn scored
0.397, the point estimate of the in-domain controller. Fig.~\ref{fig:pitvis_example} shows the transferred controller
on a PitVis test procedure. It refers to the ring curette and the tumour-excision phase, entities that do not exist in
cholecystectomy, through the same pointers, and accumulates evidence exactly as on Cholec80.

An exploratory analysis of adapting perception from a few annotated target videos is reported in
\ref{app:fewshot}.

\subsection{Failure modes}\label{sec:res_failure}
\begin{figure*}[!tp]\centering
\includegraphics[width=\textwidth]{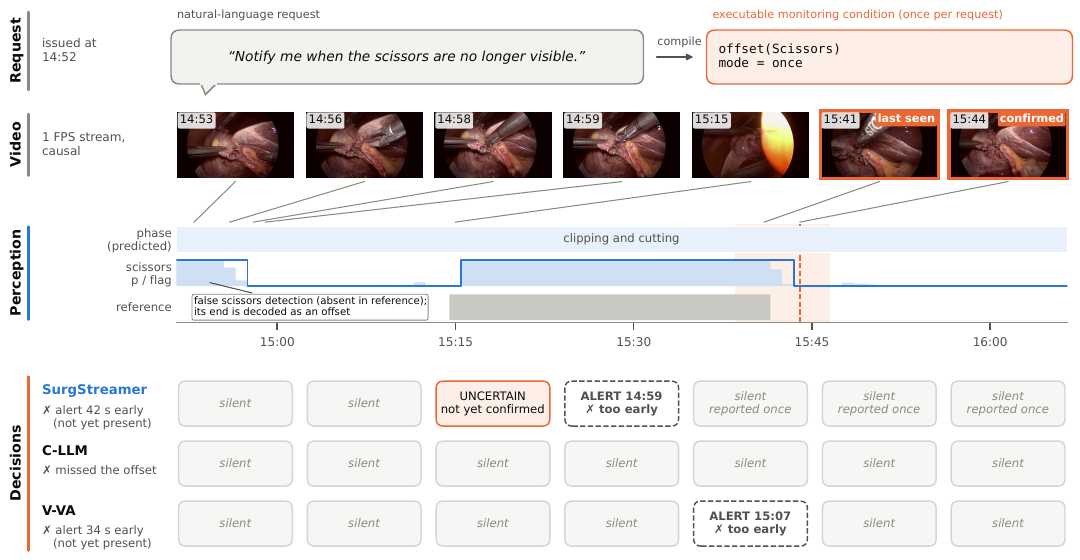}
\caption{A failure on a sealed Cholec80 test window (equal context). At the window start, perception decodes the
scissors as present although the reference marks them absent. When this false detection ends (14:58), the decoded
state contains an offset, and C-learn alerts at 14:59, 42\,s before the reference disappearance; in once mode, the real
disappearance is then not reported. C-LLM stays silent; V-VA alerts 34\,s early, and V-LoRA (not shown) also too
early. The error originates in
perception: on the decoded state, an offset had occurred. Layout as in Fig.~\ref{fig:teaser}.}
\label{fig:failure}
\end{figure*}

SurgStreamer errs where the decoded state departs from the reference. Fig.~\ref{fig:failure} shows an example from the
sealed test. At the start of a window, perception decodes the scissors as present although the reference marks them
absent; when this false detection ends, the decoded state contains a disappearance, and C-learn alerts 42\,s before
the real one, which, in once mode, is then not reported. The same mechanism underlies the fragmented episodes at
the external centres (Section~\ref{sec:res_heichole}) and the alerts on persistent presence
(Section~\ref{sec:res_vlm}).

\FloatBarrier
\subsection{Supporting analyses}\label{sec:res_dev}

\begin{table*}[!tp]\centering\small
\caption{Controllers on the full-procedure benchmark: macro event-F1 over the 13 proactive families (mean $\pm$ s.d. over 3 controller seeds where applicable). CV: out-of-fold over the training procedures (C-learn: controller-level 5-fold CV; hybrid: nested selection); dev test: 21 development procedures (no selection); final: the 12 sealed procedures, opened once; final, held-out comp.: final-test requests whose entity composition (tool pair or instrument--verb pair) never occurs in training (co-occurrence, onset-after and instrument--verb families). The hybrid choices used on dev / final test were fixed on CV.}\label{tab:controllers}
\begin{adjustbox}{max width=\linewidth}
\begin{tabular}{lcccc}\toprule
Controller & CV & Dev test & Final test & Final, held-out comp. \\\midrule
C-rule, 256-s state & 0.546 & 0.526 & 0.562 & 0.401 \\
C-rule, 1024-s state & 0.545 & 0.529 & 0.570 & 0.477 \\
P-rule (rules on semi-Markov episode posteriors), 256-s state & 0.419 & -- & -- & -- \\
C-learn, 256-s state & 0.566 $\pm$ 0.008 & 0.556 $\pm$ 0.012 & 0.591 $\pm$ 0.009 & 0.424 $\pm$ 0.029 \\
C-learn, 1024-s state & 0.567 $\pm$ 0.003 & 0.563 $\pm$ 0.005 & 0.607 $\pm$ 0.005 & 0.464 $\pm$ 0.041 \\
\textbf{Hybrid} (per-family state $\times$ controller) & 0.579 $\pm$ 0.003 & 0.563 $\pm$ 0.004 & 0.598 $\pm$ 0.007 & 0.436 $\pm$ 0.005 \\
\bottomrule\end{tabular}
\end{adjustbox}\end{table*}

On cross-validation, C-learn outperformed the exact rules (0.566 vs 0.546; Table~\ref{tab:controllers}). The gains
were on temporal and relational operators: onset during a phase 0.69 vs 0.54, $n$-th occurrence 0.51 vs 0.41. A rule
evaluator that read calibrated posteriors of an explicit-duration (semi-Markov) episode model instead of decoded
flags (P-rule) was clearly worse (0.419). The
controller's gain is therefore not explained by access to confidences alone. Perception development, the
annotation-agreement reference and the compiler analysis are reported in \ref{app:results}. In brief:
fine-tuning the encoder was the largest single perception gain, and the compiler, revised on the development request
set, reproduced 91.5\,\% of those requests exactly. Compiled instead of gold conditions lowered macro event-F1 by
0.016 on development data; the compiled system was not evaluated on the final test.

\section{Discussion}\label{sec:discussion}
\paragraph{Why does state grounding help?} The requests that matter are about events: appearances, disappearances,
phase transitions and their composition. SurgStreamer gives them exact semantics over a state stream. Given oracle
state, the exact rules answer every benchmark request correctly by construction, and the learned controller nearly so
(0.947 on PitVis oracle state). On predicted state, the results therefore suggest that perception and event timing are
the main limitation (Sections~\ref{sec:res_heichole}--\ref{sec:res_failure}). The tested VLM configurations detected appearances far better
than disappearances or phase entries. Giving them SurgStreamer's state as text helped substantially, yet a
4B-parameter model reading that state remained below a 64-unit controller reading it through pointers. We hypothesise
that the difference lies in the representation of time: the controller receives explicit run lengths and times since
events, whereas the VLM must infer them from a 16-update dialogue. Our experiments do not isolate this mechanism. The
equal-context study shows that SurgStreamer's advantage on Cholec80 does not depend on access to the procedure history.
A further property of the design, which we did not evaluate with users, is that every decision can be traced to the
decoded state and the condition that produced it (Figs.~\ref{fig:teaser}, \ref{fig:offset} and \ref{fig:failure}), so
that an error can be attributed to perception or to the reasoning by inspection; an end-to-end VLM does not expose
such an intermediate state.

\paragraph{What transfers?} We distinguish three levels of reuse: the rule semantics, the controller weights and the
controller's alert thresholds. The rule semantics transfer by construction. On oracle state, the controller weights
trained on cholecystectomy transferred unchanged to a procedure with a different anatomy, camera and vocabulary, and
performed like a controller trained there. On predicted state, unchanged weights and thresholds did not meet the
pre-registered non-inferiority margin, whereas re-calibrating only the thresholds on three validation procedures
reached the in-domain point estimate. Perception did not transfer across centres at the 2-s event level, but
re-training only perception doubled performance without changing the reasoning. The adaptation boundary of the design
therefore held for the rule semantics and the controller weights, while the thresholds belong to the
procedure-specific side.

\paragraph{What limits performance on real input?} On external data, the diagnostics implicate instrument appearance,
episode fragmentation and annotation conventions. The failure cases (Section~\ref{sec:res_failure}) show how such
errors reach the decision: a condition evaluated correctly on a wrong decoded state still produces a wrong alert. On HeiChole, SurgStreamer scored nominally higher with equal context
than with full history (0.363 vs 0.324, n.s.); one possible explanation, which we did not test, is that errors of the
frozen perception accumulate over long streams. The raw-video V-LoRA, built on a general-purpose visual encoder, was
nominally better on HeiChole, which suggests that surgical fine-tuning may trade robustness for in-domain accuracy.

\paragraph{Limitations.} The reasoning was trained on one procedure, and transfer was tested on one new procedure and
one external dataset with small test sets (7 and 9 procedures). Event timing is bounded by annotation conventions,
and rare actions and unseen entity compositions remain hard; which part of the controller (recurrence or slot
features) produces its gain was not isolated. The VLM conclusions hold for the tested OneStreamer-4B configurations,
each run once with sampling. Requests were generated from templates over a closed vocabulary, the main results use
gold conditions, and in-vocabulary objects that are simply not visible were not tested. No surgeon wrote requests
or rated alerts, and no prospective evaluation was performed, so clinical utility remains to be shown.

\paragraph{Future work.} Natural next steps are free-text requests written by surgeons, a failure attribution across
compiler, perception, decoding and control, and a sensitivity analysis of the episode parameters $G$ and $K$.
Wider baselines are also needed, including other streaming models, a state-of-the-art recognition model combined with
C-rule, and a retrospective baseline for the event memory. Because SurgStreamer shares its response space, it can be
evaluated directly on the proactive items of MedStreamBench \citep{wang2026medstreambench} drawn from the procedures
studied here, which we have not done. Multi-procedure surgical pretraining and label-free
adaptation of perception are promising ways to reduce the annotation needed per centre.

\FloatBarrier
\section{Conclusion}\label{sec:conclusion}
SurgStreamer answers natural-language monitoring requests over surgical video streams by evaluating compiled conditions
over a structured state, with learned or exact reasoning and an event memory. On a pre-registered sealed test with
gold conditions, it outperformed the tested OneStreamer-4B configurations under equal context, including fine-tuned
and vision-adapted ones, at a fraction of their cost. Its controller weights transferred unchanged to a new procedure
on oracle state; on predicted state, they matched an in-domain controller only after re-calibrating the alert
thresholds. Its perception had to be re-trained for new centres and procedures. These results support state-grounded,
modular designs for proactive surgical monitoring and point to perception under distribution shift as the main
remaining obstacle.

\section*{CRediT authorship contribution statement}
\textbf{Bowen Xiang:} Conceptualization, Methodology, Software, Validation, Formal analysis, Investigation, Data
curation, Writing -- original draft, Writing -- review \& editing, Visualization. \textbf{Jon S. Heiselman:}
Writing -- review \& editing, Visualization.

\section*{Ethics statement}
This study used only publicly available, de-identified surgical video datasets (Cholec80, CholecT50, HeiChole and
PitVis) under their respective licences. No new data were collected. The graphical abstract and
Figs.~\ref{fig:teaser}, \ref{fig:overview}, \ref{fig:memory}, \ref{fig:offset}, \ref{fig:pitvis_example} and
\ref{fig:failure} reproduce de-identified endoscopic frames from Cholec80 and PitVis with attribution.

\section*{Funding}
This work was supported by Vanderbilt University's Innovation Catalyst Fund. The content is solely the responsibility of
the authors and does not necessarily represent the official views of Vanderbilt University.

\section*{Declaration of competing interest}
The authors declare that they have no known competing financial interests or personal relationships that could have
appeared to influence the work reported in this paper.

\section*{Declaration of generative AI and AI-assisted technologies in the writing process}
During the preparation of this work the authors used Claude (Anthropic) to assist with writing analysis code,
generating figures and drafting text. After using this tool, the authors reviewed and edited the content as needed and
take full responsibility for the content of the published article.

\section*{Acknowledgements}
We thank the providers of the Cholec80, CholecT50, HeiChole and PitVis datasets.

\section*{Data and code availability}
All datasets are publicly available from their providers: Cholec80, CholecT50, HeiChole (TU Dresden OPARA,
DOI 10.25532/OPARA-284) and PitVis (UCL Research Data Repository). Code, benchmark definitions, splits and the
time-stamped protocol documents will be made publicly available upon publication.

\FloatBarrier
\bibliographystyle{elsarticle-harv}
\bibliography{references}

\onecolumn
\appendix
\counterwithin*{table}{section}
\counterwithin*{figure}{section}
\section{Additional results}\label{app:results}

\paragraph{Perception development.} Fine-tuning the encoder was the largest single gain (Table~\ref{tab:perception}):
the perception score rose from 0.39 to 0.47 and tool-onset F1 from 0.54 to 0.70. Long training context improved phase
transitions, and the four-head ensemble reached 0.54. The ordered phase HMM raised online phase-transition F1 from 0.29
(argmax with dwell filter) to 0.40. Rejected during development: boundary-aware losses, factorised triplet heads, a
spatial grid of patch tokens, a tool-aware phase cascade, an elapsed-time input and an explicit-duration episode model.

\begin{table}[H]\centering\small
\caption{Perception ablations on cross-validation over the 43 training procedures (out-of-fold predictions; decoders tuned by inner cross-validation). $S_\text{perc}$ is the mean of the five timing scores: episode-onset / offset F1 for actions and tools ($\pm2$\,s; macro over classes) and phase-transition F1 ($\pm10$\,s). Mean $\pm$ s.d. over seeds where several were trained.}\label{tab:perception}
\begin{adjustbox}{max width=\linewidth}
\begin{tabular}{lccccccc}\toprule
Variant & seeds & $S_\text{perc}$ & \multicolumn{2}{c}{Action on / off} & \multicolumn{2}{c}{Tool on / off} & Phase \\
\midrule
Frozen DINOv3 + temporal heads (A0) & 1 & 0.388 & 0.231 & 0.269 & 0.538 & 0.530 & 0.371 \\
Fine-tuned DINOv3 + presence heads (A0) & 1 & 0.468 & 0.296 & 0.261 & 0.698 & 0.656 & 0.430 \\
+ event heads (A1) & 2 & 0.488 $\pm$ 0.005 & 0.323 & 0.273 & 0.716 & 0.671 & 0.458 \\
+ event heads + boundary-aware loss (A2) & 1 & 0.481 & 0.300 & 0.270 & 0.712 & 0.686 & 0.436 \\
+ event heads + factorised I$\to$V$\to$T heads (A3) & 1 & 0.470 & 0.287 & 0.263 & 0.721 & 0.682 & 0.394 \\
A1 + elapsed-time input & 1 & 0.485 & 0.307 & 0.285 & 0.725 & 0.666 & 0.442 \\
A1, 1024-s context (A1L) & 3 & 0.498 $\pm$ 0.012 & 0.298 & 0.274 & 0.733 & 0.695 & 0.489 \\
A1, 2048-s context & 1 & 0.499 & 0.301 & 0.267 & 0.734 & 0.687 & 0.507 \\
A1L + spatial patch grid & 1 & 0.481 & 0.285 & 0.280 & 0.714 & 0.700 & 0.424 \\
A1L + tool-aware phase cascade & 3 & 0.493 $\pm$ 0.008 & 0.299 & 0.267 & 0.722 & 0.696 & 0.480 \\
Presence heads, 1024-s context (A0L) & 1 & 0.506 & 0.322 & 0.314 & 0.709 & 0.658 & 0.527 \\
Ensemble, 256-s context (A0, A1 $\times$2) & 1 & 0.511 & 0.346 & 0.289 & 0.744 & 0.714 & 0.463 \\
\textbf{Ensemble, 1024-s context (A1L $\times$3, A0L)} & 1 & 0.537 & 0.341 & 0.310 & 0.756 & 0.734 & 0.542 \\
\bottomrule\end{tabular}
\end{adjustbox}\end{table}

\paragraph{Annotation agreement.} Cholec80 and CholecT50 annotate instrument presence independently on 45 shared
procedures. Their agreement, expressed as episode-onset F1 at $\pm$2\,s, is 0.55--0.89 depending on the instrument
(Table~\ref{tab:noise_floor}). We use it as a cross-annotation agreement reference: it indicates how much of the
event-timing error at a 2-s tolerance can be attributed to annotation conventions rather than to the model.

\begin{table}[H]\centering\small
\caption{Agreement between the two independent instrument-presence annotations (Cholec80 vs CholecT50) on 45 shared procedures, as episode-onset / offset F1. A cross-annotation agreement reference for event timing, not an upper bound.}\label{tab:noise_floor}
\begin{adjustbox}{max width=\linewidth}
\begin{tabular}{lccc}\toprule
Instrument & Onset F1 ($\pm2$\,s) & Onset F1 ($\pm5$\,s) & Offset F1 ($\pm2$\,s) \\\midrule
Grasper & 0.55 & 0.61 & 0.53 \\
Bipolar & 0.81 & 0.83 & 0.77 \\
Hook & 0.80 & 0.83 & 0.79 \\
Scissors & 0.89 & 0.91 & 0.89 \\
Clipper & 0.85 & 0.88 & 0.76 \\
Irrigator & 0.82 & 0.83 & 0.79 \\
\bottomrule\end{tabular}
\end{adjustbox}\end{table}

\paragraph{Compiler.} The compiler reproduced 91.5\,\% of the unique requests of the development request set exactly
and mapped none of the ten out-of-vocabulary requests to a real instrument; a plain prompt with schema validation
mapped eight of them (Table~\ref{tab:compiler}). The compiler was revised after inspecting its errors on this request
set, so these numbers are development estimates, not held-out performance.

\begin{table}[H]\centering\small
\caption{Request compiler on the 529 unique requests of the development request set (the compiler was revised after inspecting its errors on this set, so these are development estimates): exact match of the compiled condition, split by seen / held-out templates; unsafe = an out-of-vocabulary object compiled to a real instrument; rejections = in-vocabulary requests compiled to an unsupported entity (the monitor stays silent).}\label{tab:compiler}
\begin{adjustbox}{max width=\linewidth}
\begin{tabular}{lccccc}\toprule
Compiler & Exact & Seen tpl. & Held-out tpl. & Unsafe & Rejections \\\midrule
Prompt + schema validation & 0.858 & 0.846 & 0.873 & 8 / 10 & 11 \\
+ named vocabulary, rules, normalisation, lexical grounding & 0.915 & 0.944 & 0.881 & 0 / 10 & 14 \\
\bottomrule\end{tabular}
\end{adjustbox}\end{table}

\paragraph{Tolerance.} Table~\ref{tab:tolerance_windows} repeats the equal-context window comparison at matching
tolerances from 2 to 20\,s. SurgStreamer scores highest at every tolerance; the order among the baselines changes.

\begin{table}[H]\centering\small
\caption{Sensitivity of the window comparison (Cholec80 final test) to the matching tolerance $\delta$ (same tolerance for all families), equal context. SurgStreamer scores highest at every tolerance; the order among the baselines changes.}\label{tab:tolerance_windows}
\begin{adjustbox}{max width=\linewidth}
\begin{tabular}{lcccc}\toprule
Method & $\delta=2$ & $\delta=5$ & $\delta=10$ & $\delta=20$ \\\midrule
SurgStreamer (equal context) & 0.623 & 0.662 & 0.735 & 0.821 \\
C-LLM (equal context) & 0.480 & 0.538 & 0.580 & 0.634 \\
V-VA & 0.473 & 0.514 & 0.575 & 0.645 \\
P-SG (equal context) & 0.388 & 0.495 & 0.555 & 0.620 \\
V-LoRA & 0.377 & 0.407 & 0.446 & 0.508 \\
VP-SG (equal context) & 0.284 & 0.346 & 0.409 & 0.535 \\
V (zero-shot) & 0.085 & 0.123 & 0.176 & 0.278 \\
\bottomrule\end{tabular}
\end{adjustbox}\end{table}

\section{Few-shot adaptation (exploratory)}\label{app:fewshot}

\paragraph{Design.} The study asks how many annotated target videos are needed to adapt perception when the reasoning
is kept fixed. For $k\in\{1,3,5,10\}$, five random draws of $k$ annotated target procedures train a light temporal head
on a frozen backbone; the C-learn weights trained on Cholec80 are reused unchanged. All component choices were made
on development pools disjoint from the test procedures and fixed in the protocol of 2026-10-09
(Table~\ref{tab:prereg}). The PitVis and HeiChole test procedures had been inspected in the earlier transfer studies,
so the test curves are exploratory.

\begin{figure}[H]\centering
\includegraphics[width=\linewidth]{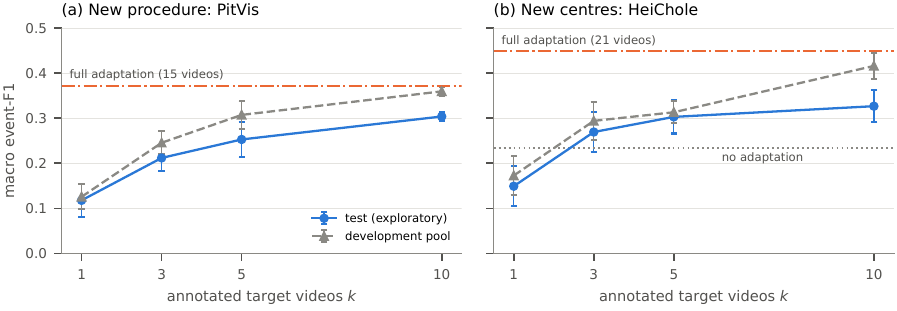}
\caption{Few-shot adaptation with frozen controller weights (exploratory): macro event-F1 versus the number of
annotated target videos used to train a light temporal head on the frozen Cholec80-tuned backbone (mean $\pm$ s.d.
over 5 draws). Solid: test procedures; dashed: development pool. Dash-dotted: full adaptation; dotted: no adaptation.}
\label{fig:fewshot}
\end{figure}

\paragraph{Results.} Performance increased with the number of annotated videos on both targets (Fig.~\ref{fig:fewshot},
Table~\ref{tab:fewshot}), and the development curves show the same pattern. On HeiChole, a single annotated video was
worse than no adaptation (0.149 vs 0.235), and three videos were on par with it. With ten videos, both targets stayed
below full adaptation, which re-trains perception, including backbone fine-tuning, on all training procedures (PitVis
0.304 vs 0.372; HeiChole 0.327 vs 0.449). Full adaptation also uses more videos (15 and 21), so the comparison mixes
the number of videos with the adaptation method.

\begin{table}[H]\centering\small
\caption{Few-shot adaptation (exploratory): macro event-F1 (mean $\pm$ s.d. over 5 draws of $k$ annotated target procedures; light temporal head on the frozen Cholec80-tuned backbone; C-learn weights trained on Cholec80, unchanged). $k=0$: unadapted Cholec80 perception (HeiChole only). Full: perception re-trained on all training procedures with backbone fine-tuning (same test procedures). Test sets had been inspected before; their results are exploratory.}\label{tab:fewshot}
\begin{adjustbox}{max width=\linewidth}
\begin{tabular}{lcccccc}\toprule
Target, evaluation & $k=0$ & $k=1$ & $k=3$ & $k=5$ & $k=10$ & Full \\\midrule
PitVis, test (exploratory) & -- & 0.118 $\pm$ 0.036 & 0.212 $\pm$ 0.028 & 0.253 $\pm$ 0.039 & 0.304 $\pm$ 0.009 & 0.372 \\
PitVis, development pool & -- & 0.126 $\pm$ 0.028 & 0.246 $\pm$ 0.026 & 0.307 $\pm$ 0.031 & 0.360 $\pm$ 0.010 & -- \\
HeiChole, test (exploratory) & 0.235 & 0.149 $\pm$ 0.045 & 0.269 $\pm$ 0.043 & 0.303 $\pm$ 0.037 & 0.327 $\pm$ 0.036 & 0.449 \\
HeiChole, development pool & 0.218 & 0.173 $\pm$ 0.043 & 0.294 $\pm$ 0.042 & 0.313 $\pm$ 0.024 & 0.416 $\pm$ 0.028 & -- \\
\bottomrule\end{tabular}
\end{adjustbox}\end{table}

\paragraph{Component decisions.} The pre-registered decisions favoured simplicity. The backbone fine-tuned on
cholecystectomy transferred better than generic DINOv3, also to pituitary surgery (0.29 vs 0.19 at $k=5$ on the
development pools), and the frozen learned controller exceeded the exact rules at every $k$ on both targets. Neither
candidate add-on met its adoption rule. Calibrating the adapter lowered false alerts by 35--40\,\% but did not raise
macro event-F1, and a controller trained on simulated perception errors lost 0.035 on Cholec80 and did not improve
transfer.

\section{Splits}\label{app:splits}
Cholec80 / CholecT50 procedure identifiers (CholecT50-only procedures are 92, 96, 103, 110 and 111).
Train (43): 1, 2, 3, 4, 5, 9, 11, 13, 15, 17, 19, 20, 21, 24, 25, 26, 27, 28, 30, 31, 33, 34, 36, 38, 43, 48, 49, 52,
56, 57, 60, 62, 67, 68, 70, 71, 72, 75, 79, 92, 96, 103, 110.
Validation (9): 8, 12, 29, 41, 44, 45, 46, 50, 78.
Development test (21): 6, 10, 14, 32, 42, 51, 53, 54, 55, 58, 59, 61, 63, 64, 69, 73, 74, 76, 77, 80, 111.
Final test (12): 7, 16, 18, 22, 23, 35, 37, 39, 40, 47, 65, 66.
PitVis: train 2, 3, 5, 7, 8, 9, 10, 12, 15, 16, 17, 18, 22, 24, 25; validation 1, 6, 23; test 4, 11, 13, 14, 19, 20, 21.
HeiChole: official split, procedures 1--24 and 25--33; for perception adaptation, training 1--21, validation 22--24,
test 25--33.

\section{Implementation details}\label{app:impl}
The frame encoder was trained with AdamW \citep{loshchilov2019adamw}: backbone learning rate $2\cdot10^{-5}$, heads
$10^{-3}$, weight decay 0.05, 200 warm-up steps and linear decay. It used batches of 32 frames sampled every second
second, random 85--100\,\% crops, horizontal flips and brightness / contrast jitter, for 6000 steps, with $w_{\max}=20$
for tools and 50 for actions.

The temporal heads used 1024-s chunks with batches of 8 (256-s chunks with batches of 16 for the short-context
ensemble), dropout 0.2 and AdamW with learning rate $10^{-3}$. Positive weights were clipped at 50 for presence and phase entry, and
at 200 for tool and action onset / offset events. Validation ran every 300 steps, with patience 6 and at most 6000 steps.

C-learn was trained on 600-s windows, half of them containing an occurrence at a random position, with 30\,\% ground-truth state, AdamW with
learning rate $2\cdot10^{-3}$ and batches of 32. Its checkpoint was selected by the validation event-F1 macro-averaged
over operators, and its thresholds were chosen from $\{0.05,\dots,0.95\}$.

The LoRA baselines used rank 16, scaling factor 32 and dropout 0.05 on all attention and MLP projections of the language
model, effective batch 8, learning rate $10^{-4}$, and selection by a fixed set of validation updates. V-LoRA and C-LLM
were trained for 2000 steps. V-VA was trained for 4000 steps, half of them auxiliary state descriptions, with
vision-encoder LoRA and fully trained projectors (learning rate $2\cdot10^{-5}$), on one A100 GPU. All other training
and all inference used one RTX 5070 (12\,GB). The few-shot calibration that was not adopted used per-tool Platt scaling
\citep{platt1999} and one shared hysteresis setting, fitted on cross-fitted predictions of the $k$ adaptation
procedures.

\end{document}